%% file: arxiv.tex
\documentclass[conference]{IEEEtran}
\IEEEoverridecommandlockouts
\DeclareRobustCommand*{\IEEEauthorrefmark}[1]{%
  \raisebox{0pt}[0pt][0pt]{\textsuperscript{\footnotesize #1}}%
}

\input{latex_headers}

\usepackage[misc]{ifsym}

\title{Modeling Global Dynamics from Local Snapshots with Deep Generative Neural Networks}

\author{\IEEEauthorblockN{Scott Gigante\,\IEEEauthorrefmark{1}\,$^\dag$,
David van Dijk\,\IEEEauthorrefmark{2}\,\IEEEauthorrefmark{3}\,$^\dag$,\thanks{$^\dag$ These authors contributed equally. $^{\dag\dag}$ These authors contributed equally.}
Kevin R. Moon\IEEEauthorrefmark{4}, \\
Alexander Strzalkowski\IEEEauthorrefmark{5},
Guy Wolf\,\IEEEauthorrefmark{6}\,$^{\dag\dag}$\, and
Smita Krishnaswamy\,\IEEEauthorrefmark{2}\,\IEEEauthorrefmark{3}\,$^{\dag\dag}$ $^{\S}$\, \thanks{$^\S$ Corresponding author. \texttt{smita.krishnaswamy@yale.edu} \newline \indent\indent~ 333 Cedar St. New Haven CT 06510 USA}
} 
\IEEEauthorblockA{\IEEEauthorrefmark{1} Computational Biology and Bioinformatics Program,
\IEEEauthorrefmark{2} Department of Genetics, \\
\IEEEauthorrefmark{3} Department of Computer Science, Yale University, New Haven, CT, USA}
\IEEEauthorblockA{\IEEEauthorrefmark{4} Department of Mathematics and Statistics, Utah State University, Logan, UT, USA}
\IEEEauthorblockA{\IEEEauthorrefmark{5} Department of Computer Science, Princeton University, Princeton, NJ, USA}
\IEEEauthorblockA{\IEEEauthorrefmark{6} Department of Mathematics and Statistics, Universit\'{e} de Montr\'{e}al, Mont\'{e}al, QC, Canada}
}

\begin{document}

\maketitle

\begin{abstract}
Complex high dimensional stochastic dynamic systems arise in many applications in the natural sciences and especially biology. However, while these systems are difficult to describe analytically, ``snapshot'' measurements that sample the output of the system are often available. In order to model the dynamics of such systems given snapshot data, or local transitions, we present a deep neural network framework we call \textit{Dynamics Modeling Network} or \textit{DyMoN}. DyMoN is a neural network framework trained as a deep generative Markov model whose next state is a probability distribution based on the current state. DyMoN is trained using samples of current and next-state pairs, and thus does not require longitudinal measurements. We show the advantage of DyMoN over shallow models such as Kalman filters and hidden Markov models, and other deep models such as recurrent neural networks in its ability to embody the dynamics (which can be studied via perturbation of the neural network) and generate longitudinal hypothetical trajectories. We perform three case studies in which we apply DyMoN to different types of biological systems and extract features of the dynamics in each case by examining the learned model.
\end{abstract}

\section{Introduction}
\label{introduction}

\input{introduction}

\section{Dynamics Modeling Network (DyMoN)}

Let $\mathcal{X} \subseteq \mathbbm{R}^d$ be a finite dataset of $d$-dimensional states, and let $\mathcal{T} \subseteq \mathcal{X} \times \mathcal{X}$ be a (finite) collection of transitions between these states. Such transitions can either be observed by sampling a dynamical system, or constructed by geometry-revealing diffusion methods such as diffusion maps~\cite{coifman2006diffusion}. We propose to learn the dynamics in $\mathcal{T}$ and the geometry represented by them as a stochastic velocity vector field of a Markov process with a feed-forward neural network we call Dynamics Modeling Network (DyMoN).

DyMoN is formed by a cascade of linear operations and nonlinear activations. These are controlled by optimized network weights, which we collectively denote by $\theta \in \Theta$, where $\Theta$ represents the space of possible weight values determined by the fixed network architecture. To capture the dynamics in $\mathcal{T}$, DyMoN learns a \emph{velocity vector} $\Delta_{\theta}(x) \in \mathbbm{R}^d$ and uses it to define a transition function $T : \mathbbm{R}^d \times \Theta \to \mathbbm{R}^d$ that generates a Markov process
$$
x_t = T(x_{t-1},\theta) = x_{t-1} + \Delta_{\theta}(x_{t-1}) \qquad t = 1,2,3,\ldots,
$$
given an initial state $x_0 \in \mathcal{X}$ and optimized weights $\theta \in \Theta$. Further, to introduce stochasticity, we treat the output $\Delta_{\theta}(x_{t-1})$ as a random vector whose probability distribution determines the conditional probabilities $P(x_{t} | x_{t-1})$ of the Markov process. 

\begin{figure}
\caption{(\textbf{A}) Schema of DyMoN architecture where $x$ represents the neural network input vector, $\hat{y}$ represents the predicted network output in (\ref{eq:output}), and $\epsilon$ is a random Gaussian noise vector. $P(y|x)$ represents the distribution of outputs from many samples of the stochastic dynamical process given $x$, and $P(\hat{y}|x)$ represents the distribution of outputs of DyMoN given $x$ and many different noise vectors. (\textbf{B}) Example input states. (\textbf{C}) Learned transition vectors (arrows) and output states from DyMoN.}
\label{fig:schema}
\includegraphics[width=\linewidth]{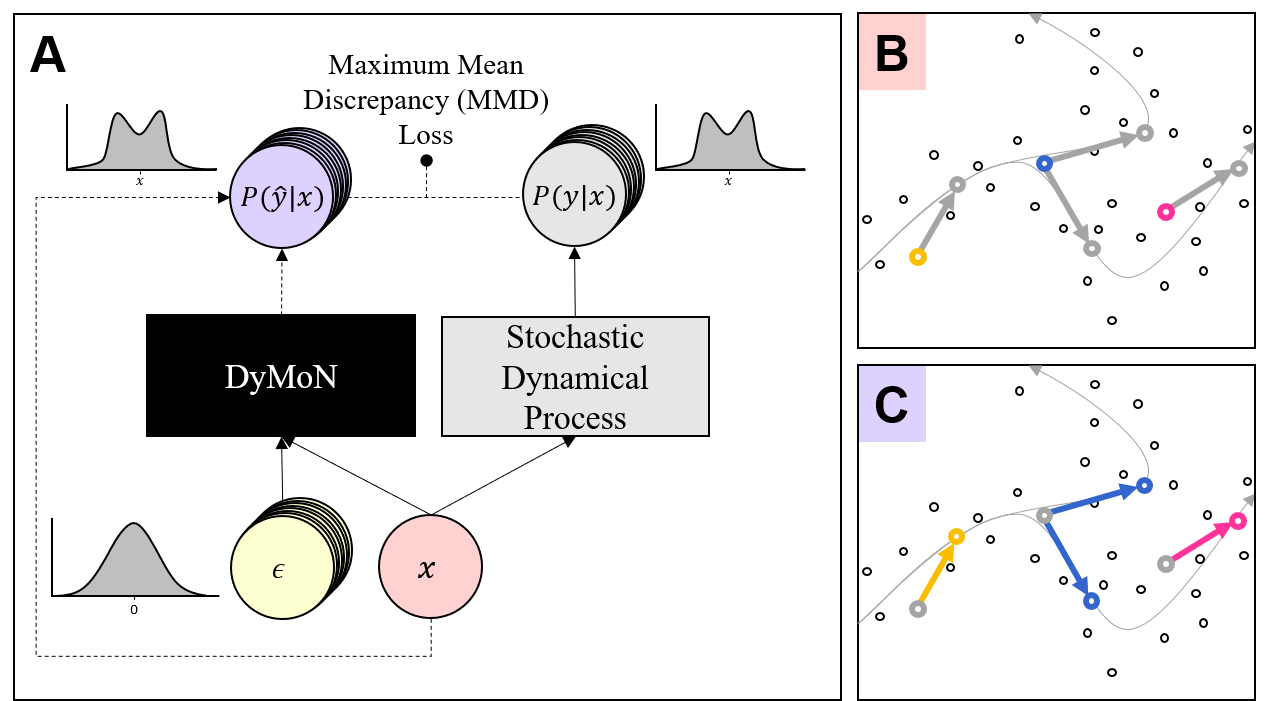}
\end{figure}

Given a fixed architecture, we now describe how DyMoN is trained using the data points in $\mathcal{X}$. Consider a source state $x \in \mathcal{X}$ with multiple transitions from it leading to target states $Y_x = \{y : (x,y) \in \mathcal{T}\}$. Let $\mathcal{P}_x(y)=P(y \mid x)$, whose support is $Y_x$, be the conditional probability of transitioning from $x$ to $y$. For ease of notation, we allow repetitions in the set notation of $Y_x$, and assume that data points in $Y_x$ are indeed distributed according to $\mathcal{P}_x(y)$ and are sufficient for estimating $\mathcal{P}_x(y)$. Further, we assume that these transitions are smooth in the sense that if $x$ is similar to $x^\prime$ then $P(y \mid x)$ is similar to $P(y \mid x^\prime)$. Therefore in practice we may replace $Y_x$ in the following training procedure with $\cup_{x^\prime \approx x} Y_{x^\prime}$, which includes target points from neighbors of $x$  to increase the robustness of DyMoN.

The stochastic output of DyMoN is enabled by a random input vector $\varepsilon \in \mathbbm{R}^n$ sampled from a probability distribution $\mathcal{F}$ with zero mean and unit variance. This input can be explicitly written into DyMoN function as
\begin{equation}
T_\varepsilon(x,\theta) = x + f_{\theta}(x,\varepsilon),
\label{eq:output}
\end{equation}
where $f_{\theta} : \mathbbm{R}^d \times \mathbbm{R}^n \to \mathbbm{R}^d $ represents suitable feed forward layers for combining the training input $x$ with the random input $\varepsilon$ to provide an instantiation of the transition velocity vector $\Delta_{\theta}(x)$. 

Given $x \in \mathcal{X}$ and $\theta \in \Theta$, one can consider the distribution $\widehat{\mathcal{P}}_x^{(\theta)}$ of the random variable $T_\varepsilon(x,\theta)$, $\varepsilon \sim \mathcal{F}$ and estimate it by $m$ i.i.d.\ instantiations $\boldsymbol{\epsilon} = \{\varepsilon_j \sim \mathcal{F}\}_{j=1}^m$ passed through the network to form $\hat{Y}^{(\boldsymbol{\epsilon})}_{(x,\theta)} = \{T_{\varepsilon_1}(x,\theta), \ldots, T_{\varepsilon_m}(x,\theta)\}$. To conform with the training data, DyMoN is optimized so that this distribution approximates the distribution $\mathcal{P}_x$, as captured from the training data by $Y_x$. This optimization is given by
$\argmin_{\theta} \mathbbm{E}\left[\{H_{x}(\theta) : x \in \mathcal{X}\right]$
with $H_{x}(\theta) = \MMD(\widehat{\mathcal{P}}_x^{(\theta)},\mathcal{P}_x)$.
 The MMD is computed using a Gaussian kernel over $\hat{Y}^{(\boldsymbol{\epsilon})}_{(x,\theta)}$ and $Y_x$. Given trained weights $\theta \in \Theta$ and an initial state $x_0 \in \mathcal{X}$, a random walk is generated by $x_{t} = T_{\varepsilon_t}(x_{t-1},\theta)$, where $t=1,2,\ldots$, and $\varepsilon_1,\varepsilon_2, \ldots \stackrel{\text{i.i.d.}}{\sim} \mathcal{F}.$
 
The training and application of DyMoN to learn deterministic systems and higher-order Markov processes extend naturally from the stochastic memoryless (i.e., first-order) DyMoN, a) by ignoring the random input; and b) providing additional previous states as input respectively. Training details, mathematical background and further evaluation of DyMoN can be found in the appendix.

\section{Empirical Validation}

In this section we demonstrate the performance and accuracy of DyMoN and compare it to a range of other methods. We show that we can reliably use DyMoN to \begin{enumerate*} \item learn the transition probabilities at each state; \item generate stochastic trajectories within a system; and \item visualize the data.
\end{enumerate*}

\textbf{Trajectory generation:} We demonstrate the ability of DyMoN to generate paths on a single and double pendulum, providing one and two previous states as input for the single and double pendulums respectively.

\begin{figure}[ht]
    \centering
    \includegraphics[width=\linewidth]{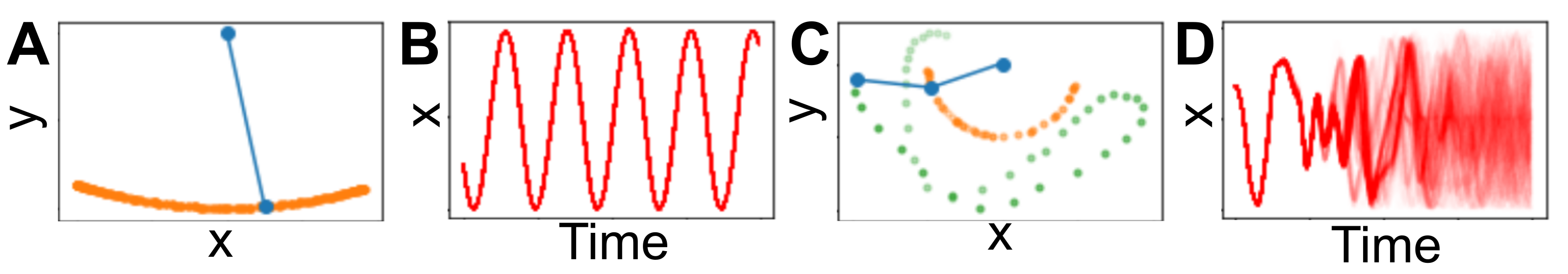}
    \caption{Paths generated by the second-order DyMoN trained on a single and double pendulum. A single path is shown for each system in (\textbf{A}) and (\textbf{C}). (\textbf{B}) and (\textbf{D}) show the x coordinate of 500 paths starting from an epsilon-difference for the single and the lower of the double pendulums, respectively.}
    \label{fig:pendulums}
\end{figure}

Figure~\ref{fig:pendulums} shows the Euclidean coordinates of DyMoN-generated paths of both pendulums over time, with only the lower pendulum shown in the case of the double pendulum. Both predicted pendulums show smooth trajectories; the single pendulum shows periodic behavior and the double pendulum shows chaotic behavior. 



\begin{figure}[ht]
    \centering
    \includegraphics[width=\linewidth]{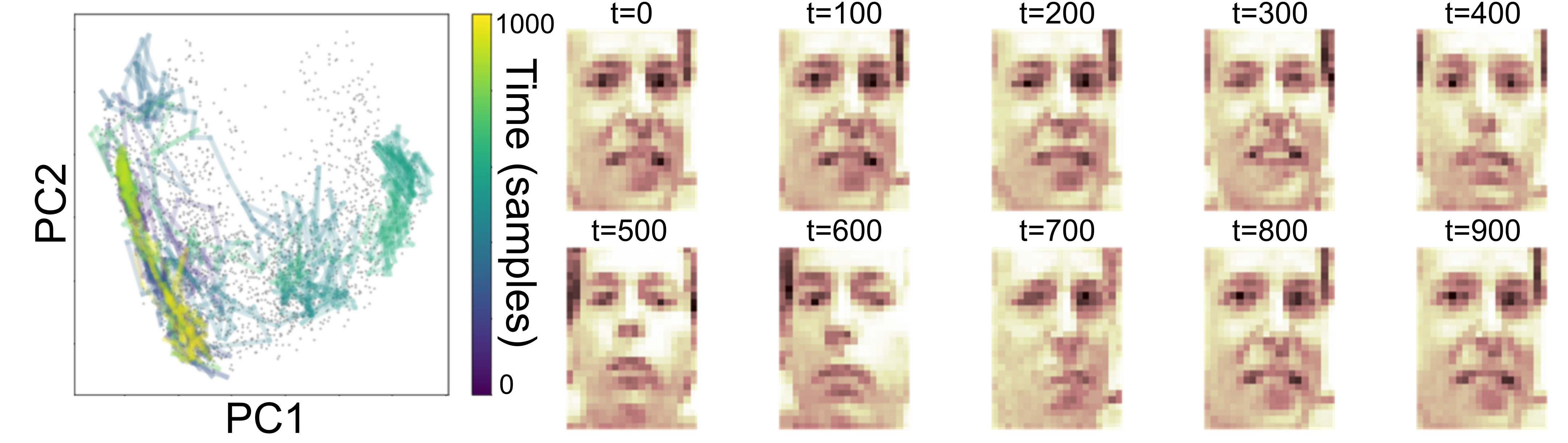}
    \caption{Chain generated by the  DyMoN visualized on a PCA embedding of the Frey faces dataset (left). The chain is shown in color (indicating time) superimposed over the training data in gray. 10 equally spaced faces generated by DyMoN are also shown (right).}
    \label{fig:face_path}
\end{figure}

We also trained DyMoN on the Frey faces dataset~\cite{roweis2000nonlinear} to demonstrate DyMoN's capacity to learn an empirical model of a stochastic system for which no generating distribution exists. Figure~\ref{fig:face_path} shows 1000 samples generated by DyMoN on the Frey faces dataset visualized using PCA and ten samples from this trajectory with uniform spacing in time. DyMoN samples a large range of states and generates a realistic new trajectory. 

\textbf{Stationary distribution comparison to other methods:} We train DyMoN to sample a Gaussian mixture model (GMM) to show it can generate samples that match the expected data distribution, avoiding the ''mode collapse`` problem which affects Generative Adversarial Networks, in which the model generates samples from only a single mode of a multimodal distribution. We generate a 1D Gaussian mixture model using Metropolis-Hastings sampling and compare the performance of Recurrent Neural Networks (RNNs), Hidden Markov Models (HMMs) and Kalman Filters (KFs) (Table~\ref{tab:gmm_performance}). DyMoN performs similarly to the HMM, both in accuracy of the sampled distribution measured by Earth Mover's Distance (EMD) and in inference time. RNNs fall into an infinite loop reproducing the same data point. Kalman filters are inherently difficult to train on a system without predefined states and transitions, and as such under-perform in an unsupervised setting.

\begin{table}
    \centering
    \caption{Performance of learning methods for dynamical systems on generating samples from a Gaussian mixture model.}
    \vskip 0.15in
    \begin{center}
    \begin{small}
    \begin{sc}
    \begin{tabular}{lccc}
    \toprule
                & EMD                     & Train & Inference \\
    \midrule
        DyMoN   & $0.150$  & 9 min     & 19.7 s \\
        RNN     & $1.61$ & 269 min   & 3209 s \\
        KF      & $0.357$  & 650 s     & 15.1 s \\
        HMM     & $0.159$  & 92 s      & 10.5 s \\
    \bottomrule
    \end{tabular}
    \end{sc}
    \end{small}
    \end{center}
    \vskip -0.1in
    \label{tab:gmm_performance}
\end{table}

\begin{figure}[ht]
	\centering
    \includegraphics[width=\linewidth]{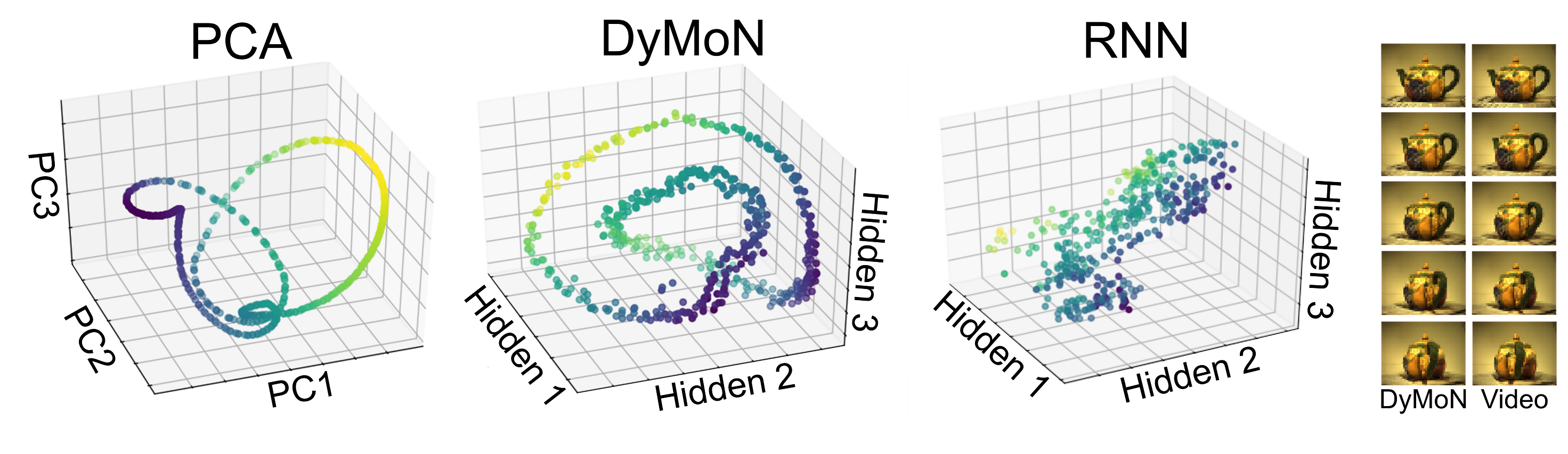}
    \caption{Embedding of the teapot video dataset colored by the x axis using PCA (left), a three-node hidden layer of DyMoN (center), and a three-node hidden layer of an RNN (right).  We also show sequential images from the original video and a DyMoN-generated chain (far right).}
    \label{fig:teapot}
\end{figure}

\textbf{Visualization:} A DyMoN with a low-dimensional latent layer can also be used to produce a visualizable embedding of the data. When trained on a video of a rotating teapot, \cite{weinberger2004learning} DyMoN's embedding layer is homeomorphic to a circle (Fig.~\ref{fig:teapot}), while PCA on the same dataset produces spurious intersections, and an RNN does not produce an interpretable embedding, as it does not simply model the transition but has memory that may confound the embedding. 

\section{Biological Case Studies}

\subsection{T Cell Development in the Thymus}

Next, we use DyMoN to learn transitions in single-cell data. 
Since single-cell measurements are destructive, we cannot follow a single cell through time with current technologies; instead, the entire population of cells is used to derive potential temporal cell trajectories~\cite{setty2016wishbone}. 

\begin{figure*}[ht]
    \centering
    \includegraphics[width=\linewidth]{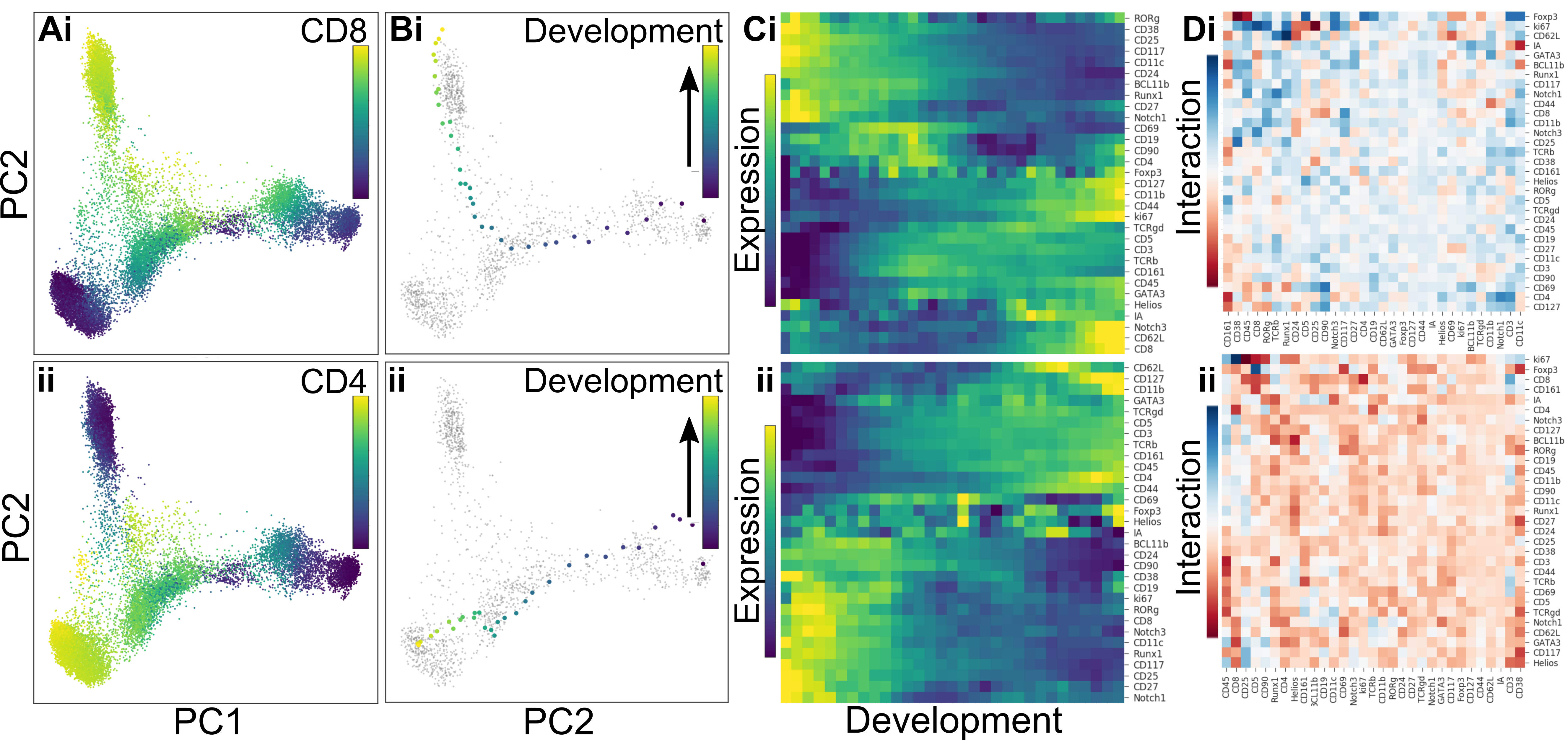}
    \caption{DyMoN on high-dimensional protein abundance data of T cell development. (\textbf{A}) PCA plots of all 17,000 cells colored by CD8 (\textbf{Ai}) and CD4 (\textbf{Aii}) expression. (\textbf{B}) PCA plots with all cells in grey and DyMoN trajectories in color. (\textbf{Bi}) DyMoN trajectory of CD4+/CD8- T helper cells. (\textbf{Bii}) DyMoN trajectory of CD4-/CD8+ cytotoxic T cells. (\textbf{C}) Shows row z-scored heatmaps of marker expression as a function of the trajectory for each of the two DyMoN trajectories with hierarchically clustered genes on the rows and cells on the columns. (\textbf{D}) Heatmaps of the Jacobians obtained at the branch point (\textbf{Di}) and at the end of the CD8+/CD4- branch (\textbf{Dii}).
    }
    \label{fig:wishbone}
\end{figure*}

Here, we use a mass cytometry dataset measuring developing T cells (adaptive immune cells) from a mouse thymus~\cite{setty2016wishbone}, using pseudotime from Wishbone~\cite{setty2016wishbone} to sample local transitions. After training, we sample trajectories by initializing DyMoN with undifferentiated cells. We obtain two types of trajectories (Fig.~\ref{fig:wishbone}.) We examine marker expression as a function of trajectory progression (Fig.~\ref{fig:wishbone}Cii and Ciii). 
The two trajectories follow the known progression of developing T cells, beginning at CD4-/CD8- immature T cells, developing into CD4+/CD8+ poised T cell progenitors, and finally diverging into T helper cells (CD4+/CD8-) and cytotoxic T cells (CD4-/CD8+) respectively.

To investigate the DyMoN learned internal representation, we compute the Jacobian of the transition vector with respect to the inputs at two points on the trajectory: at the branch point (Fig.~\ref{fig:wishbone}Di) and at the CD8+/CD4- branch (Fig.~\ref{fig:wishbone}Dii). DyMoN learns different associations between genes at each point in the ambient space. We find a negative association between CD4 and CD8 in the Jacobian obtained at the branch point, reflecting the known decision between downregulating either CD4 or CD8. In addition, we find that in the Jacobian of the CD8+ branch Gata3, a developmental marker, goes down with CD4 and CD8, confirming the developmental characteristic of the trajectory. Finally, in the same CD8+ branch Jacobian, Foxp3 and CD25 are positively associated, consistent with their role as regulatory T cell markers. Analysis of DyMon concurs with established literature on T cell development, showing that DyMoN can both learn meaningful trajectories and provide insights into the network's dynamical model of the system.

\subsection{Human Embryonic Stem Cell Differentiation}

We now apply the generative capabilities of DyMoN on a biological system of newly measured single-cell RNA sequencing data of human embryonic stem cells (hESCs) grown as embryoid bodies (EB). Cells were distributed over a 27-day time course and measured using the 10X Chromium platform. We train DyMoN on Markov transitions from the diffusion geometry of the dataset. We sample neighbors $\mathbf{y}$ of $x$ over weighted affinities defined by a Gaussian kernel, retaining neighbors for which $y$ is defined to be ``later'' than $x$ (defined from a smoothed estimate of the discrete time variable.)

\begin{figure*}
    \centering
    \includegraphics[width=0.8\textwidth]{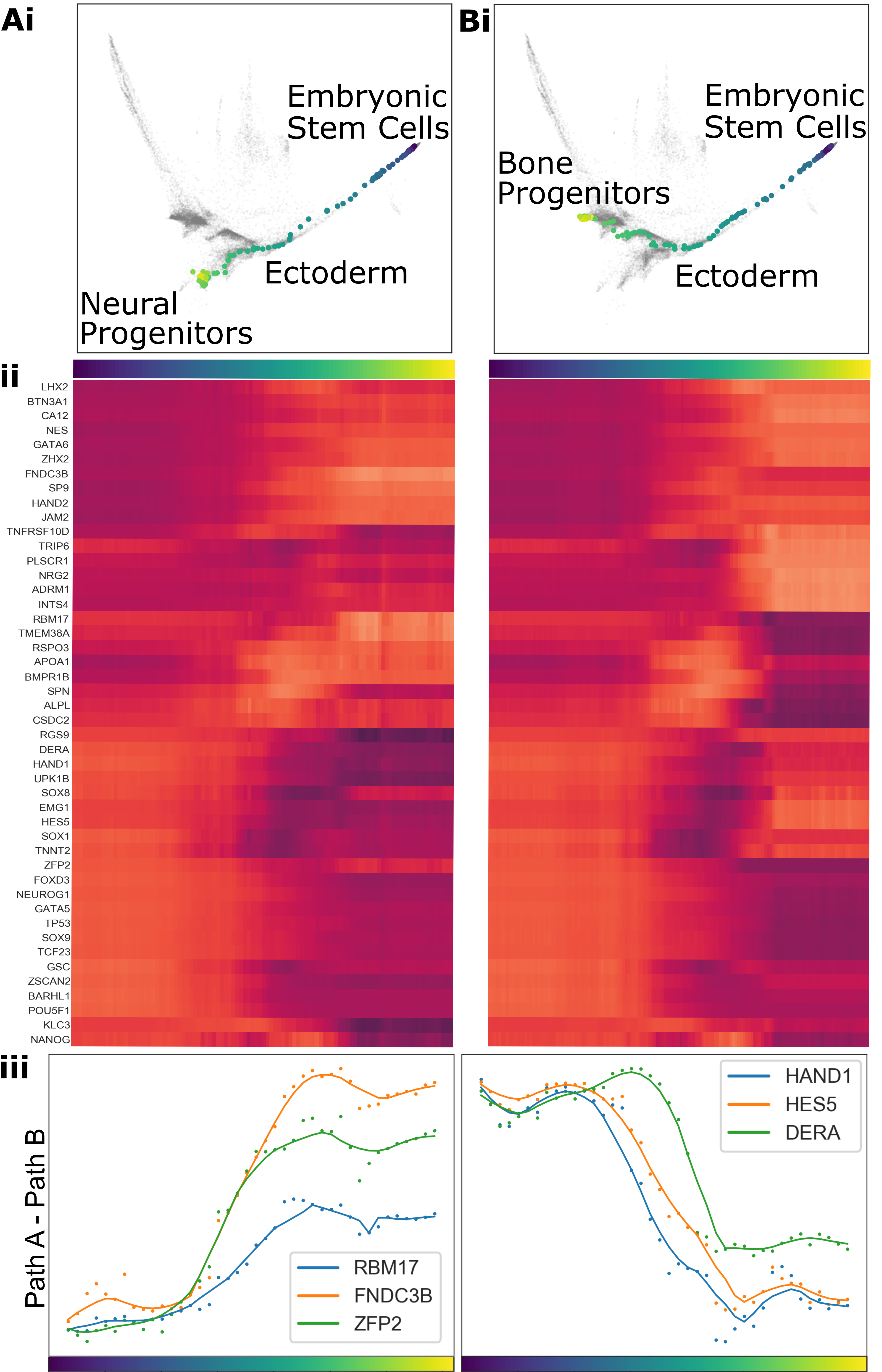}
    \caption{Paths generated by DyMoN sampling the neural progenitor (A) and bone progenitor (B) cell states. Paths are shown both on MDS (i) and as heatmaps of selected transcription factors (ii). From these paths, we propose a novel cellular programming protocol shown in (iii).}
    \label{fig:eb}
\end{figure*}

Fig.~\ref{fig:eb}, shows two generated trajectories: neural progenitor development (A) and bone progenitor development (B). Fig.~\ref{fig:eb}ii shows that each of these paths initially expresses known stem cell markers NANOG and POU5F1 and follow a common transition into the ectoderm (LHX2) before diverging, with path A taking the neural progenitor branch distinguished by SOX1, and path B taking the bone progenitor branch distinguished by ALPL. We observe a novel set of transcription factors distinguishing each stage of differentiation (Fig.~\ref{fig:eb}iii), and propose them as a potentially novel reprogramming protocol to obtain each respective mature cell type. Thus, DyMoN provides a novel form of hypothesis generation enabled by deep abstract models of dynamical systems.

\section{Conclusion}

\input{conclusion}

\subsection*{\bf{Acknowledgments}}
\noindent This research was partially funded by a grant from the Chan-Zuckerberg Initiative (ID: 182702).



\bibliographystyle{IEEEtran}
\bibliography{main}

\newpage
\input{appendix}

\end{document}

%% file: latex_headers.tex
\usepackage{microtype}
\usepackage{graphicx}
\usepackage{subfigure}
\usepackage{booktabs} 
\usepackage{floatrow}

\usepackage{amsmath}
\usepackage{amsthm}
\usepackage{amssymb}
\usepackage{bbm}
\usepackage{verbatim}
\usepackage{graphicx}
\usepackage{subfigure}
\usepackage{afterpage}
\usepackage{etoolbox}
\usepackage[inline]{enumitem}
\usepackage{float}
\usepackage{algorithmic}
\usepackage{algorithm}
\usepackage{multirow}
\usepackage{rotating}
\usepackage[dvipsnames]{xcolor}

\usepackage{siunitx}
\makeatletter
\renewcommand{\p@subfigure}{\thefigure}
\makeatother

\input{math_commands.tex}

\DeclareMathAlphabet{\mathpzc}{OT1}{pzc}{m}{it}

\DeclareMathOperator{\MMD}{MMD}

\graphicspath{{figs/}}

%% file: math_commands.tex
\usepackage{amsmath,amsfonts,bm}

\def\eqref#1{equation~\ref{#1}}

\def\1{\bm{1}}

\DeclareMathAlphabet{\mathsfit}{\encodingdefault}{\sfdefault}{m}{sl}
\SetMathAlphabet{\mathsfit}{bold}{\encodingdefault}{\sfdefault}{bx}{n}

\DeclareMathOperator*{\argmin}{arg\,min}

%% file: introduction.tex
It is difficult if not impossible to derive differential equations or analytical models for most natural stochastic dynamic systems. Further, in Biology it is often the case that we only have access to samples or ``snapshots'' from such dynamic system (e.g., pairs of consecutive points) and not to continuous observations. Here, we propose to learn a generative neural network model of dynamic systems that we call a Dynamics Modeling Network (DyMoN) that is able to learn a dynamic model of a system for which we only have sparse snapshot data, or even a single snapshot showing a multitude of instantiations of the dynamic system rather than one instantiation longitudinally. DyMoN uses a deep neural network architecture to learn fixed-memory stochastic dynamics (e.g., memoryless or $n$-th order Markov process) in an observed system. This network is trained to map a current state (or $n$ states) to a distribution of next states. DyMoN is trained by penalizing the stochastically generated output states based on maximal mean discrepancy (MMD)~\cite{gretton2012kernel,dziugaite2015mmdnet} from known next-states as a probabilistic distance between the desired and generated output distributions.

DyMoN provides several advantages over existing methods, such as HMMs and recurrent neural networks. First, DyMoN provides a \textbf{representational} advantage by serving as an embodiment of the dynamics in lieu of a predetermined model, such as stochastic differential equations. This representation is deep and factored, in the sense that the ``logic'' of the dynamic transition is broken down into increasingly abstract steps, each of which can be visualized or examined. Second, DyMoN provides a \textbf{generative} advantage as it generates new trajectories/sequences that have not been seen previously in the system. Third, the utilization of a deep network model offers a \textbf{multitasking} advantage: while previous models, such as HMMs, can simulate trajectories, and one may use PCA (or similar methods) to visualize trajectory information, most previous methods are not designed or equipped to \emph{simultaneously} learn multiscale features (i.e., in many levels of abstraction), visualize intrinsic underlying dynamics, and utilize them to generate new trajectories. Finally, the natural parallelizability of neural networks (e.g., with GPU-based implementations) offers \textbf{computational} advantages, as they can be used to process large volumes of noisy data. Furthermore, once trained, DyMoN can efficiently generate new trajectories faster than most existing methods. 

DyMoN is well suited to model biological data, such as single-cell RNA sequencing data from cellular developmental systems. Such data is often collected at only one or a handful of time-points. Thus the dynamics are either inferred from ``pseudotime'' (an inferred axis of progression based on observed cells which are at different states of development), or from interpolation between discrete time points. These trajectories are not a matter of “sequence completion” as is often done in recurrent neural networks; the goal is to predict the near-future state of a cell given its current state. Often history information is not available in biological measurements. As such, we apply DyMoN to several biological systems for which we have access to snapshot samples of the data. These include mass cytometry of developing T cells in the mouse thymus and single-cell RNA sequencing of human embryonic stem cells developing in embryoid bodies. DyMoN is able to learn the underlying dynamics of each of these systems, reveal insights into the `rules' driving these dynamics, as well as sample new trajectories/sequences from these dynamics.

%% file: conclusion.tex
Here we presented DyMoN, a neural network framework for modeling stochastic dynamics from observed samples of a system. DyMoN is well-suited for modeling $n$-th order Markovian stochastic dynamics from the types of data that occur in biological settings, especially in systems for which the generative process cannot be described by differential equations. The flexibility of this framework is enabled by several aspects of DyMoN. First, since the networks encode Markovian dynamics, they do not require continuous longitudinal data. Second, the MMD penalty used to train the next-state generation can be enforced via samples or a known probability distribution, making the networks trainable on top of shallow models such as diffusion operators and pseudotime inferences. DyMoN enables generation of new trajectories, extraction of feature dependencies, and visualization of the dynamic process, and will enable inference of driving forces (transcription factors, mutations, etc.) of progression dynamics in biomedical data.

%% file: appendix.tex
\onecolumn

\begin{appendices}

\section{Background}
\textbf{Markov Processes:} Markov processes are stochastic memoryless models that describe dynamic systems, i.e., each state depends only on the last state. However, higher order processes can be modeled by making the state dependent on several previous states. A deterministic Markov chain maps each state to only one next state whereas a stochastic Markov chain models a probabilistic transition such that at each state multiple transitions are possible. DyMoN is a neural network trained to replicate a stochastic (with input noise) or deterministic (no input noise) Markov process. 

\textbf{Maximum Mean Discrepancy:} Divergences, such as KL divergence, are used to measure distances between probability distributions. However, they require density estimation and as such divergences are hard to compute for high dimensional systems. Maximum Mean Discrepancy (MMD) \cite{gretton2012kernel} offers a solution by using the kernel trick to circumvent the curse of high dimensionality. Instead of comparing empirical distributions, MMD is defined on the inter- and intra-sample pairwise affinities, which are computed using a kernel. With MMD we can therefore compute the distributional distance (i.e. divergence) between two samples, without the need for density estimation. This distance is defined as
\begin{align*}
MMD^2(\mu,\nu) &= \iint k(x,x^{\prime}) d\mu d\mu + \iint k(y,y^{\prime})d\nu d\nu - 2 \iint k(x,y) d\mu d\nu
\end{align*}
where $k(\cdot,\cdot)$ is a kernel function and $x$ and $y$ are sampled from the two distributions represented by the probability measures $\mu$ and $\nu$. In practice, this distance is estimated using summation over two finite sets of samples.

MMD has been used in~\cite{dziugaite2015mmdnet} to translate samples from one distribution into another distribution using a deep neural network architecture. Samples were generated by sampling random values from some simple distribution, e.g. Gaussian, and running them through a deep neural network. The network was trained by minimizing the MMD between generated samples and real samples. In~\cite{shaham2017removal} this architecture was extended by using a residual network with the application of removing undesirable batch effects that are associated with measurements. We use MMD to learn a conditional distribution that represents the possible next states in the Markov chain.

\begin{figure}[t]
    \centering
    \includegraphics[width=\linewidth]{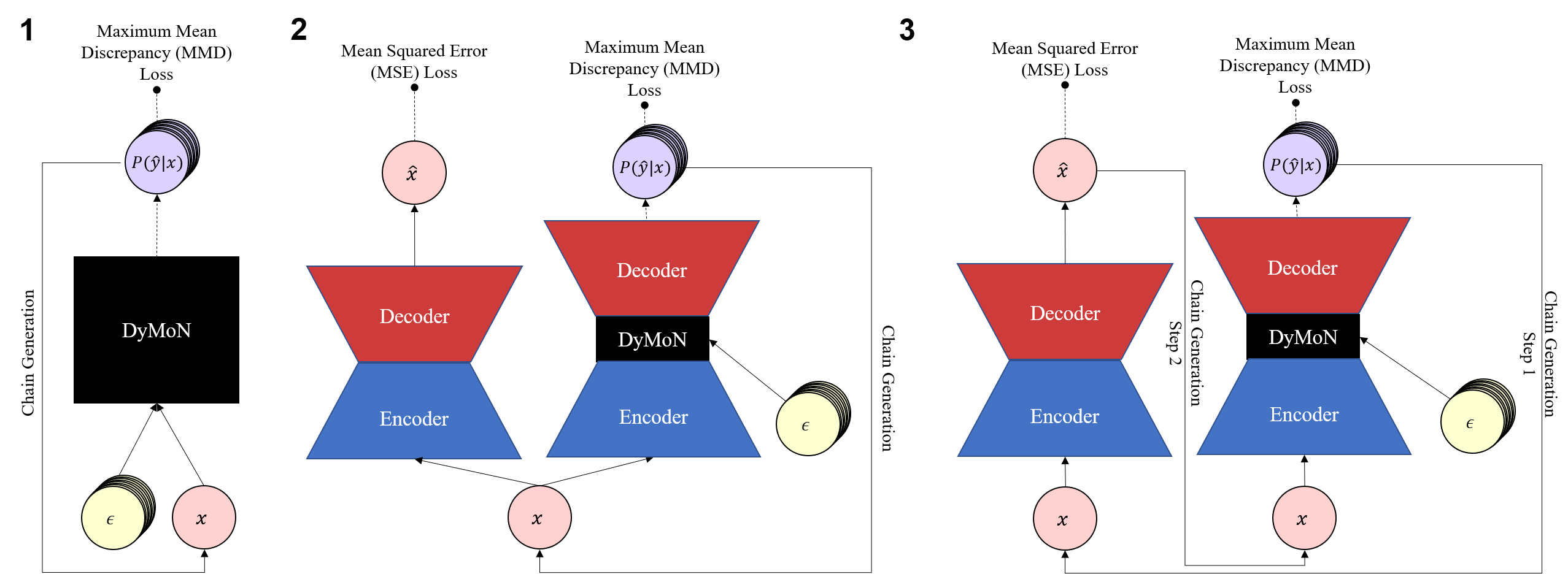}
    \caption{Alternative DyMoN architectures. The stochastic output can be generated in the ambient space (1) or the latent space of an autoencoder (2, 3). Chains of samples are generated by feeding DyMoN output back in as a new input; this can be passed additionally through the encoder and decoder as a denoising step (3).}
    \label{fig:architectures}
\end{figure}

\section{Related Work}

\textbf{Recurrent Networks (RNNs):} RNNs are trained to predict the next state of a sequence and have been used for text analysis~\cite{FernandezRNN}, speech recognition~\cite{amodei2016deepspeech}, and language modeling~\cite{Jzefowicz2016ExploringTL}, as well as other tasks that operate on time series data. In contrast to DyMoN, RNNs typically require a sequence or history to predict the next state. 

\textbf{Stochastic Generative Networks:}  There have been other networks that use stochasticity to learn conditional distributions. For example, given an input $X$ and a stochastic input to a middle layer, a variational autoencoder (VAE)~\cite{kingma2013vae} learns to transform these inputs into a Gaussian centered at a maximum likelihood point from which $X$ is derived. Thus the variational network can denoise samples and also generate samples ``like'' given samples. However, VAEs focus on generating data points instead of transitions. Several stochastic RNNs have also been proposed~\cite{bowman2016generating,chung2015recurrent}. For example, in~\cite{goyal2017variational}, a stochastic generative RNN is trained to learn a process that converges to the full data distribution within a small number of steps given a simple initialization.  In contrast, DyMoN learns the transitional probabilities from state to state of a Markov process instead of its stationary distribution. 

\textbf{Time Delay Neural Networks:} DyMoN also has some conceptual connection to Time Delay Neural Networks (TDNNs) which include a contextual window samples as input. However, the main focus of TDNNs are to classify patterns with shift-invariance, such as recognizing phenomes in speech~\cite{waibel1989phoneme}. TDNNs generally achieve this by taking a time convolution through windows of time to train a classifier. In contrast, we focus on learning Markov processes with the goal of generating plausible next-states and analyzing the dynamics that drive these transitions. 

\begin{table}[b]
    \centering
    \caption{DyMoN architecture and training details. Architecture refers to alternative DyMoN architectures shown in Figure~\ref{fig:architectures}. Where step size is given, samples are provided to DyMoN as $(x_t, x_{t+step\_size})$.}
    \vskip 0.15in
    \begin{center}
    \begin{small}
    \begin{sc}
    \begin{tabular}{lccr}
    \toprule
    Data set & Architecture & Hidden Layers & Step Size \\
    \midrule
    Pendulum & 1 & [8, 16, 8] & 1 \\
    Teapot & 2 & 2xConv, 1x3, 2xDeconv & 10 \\
    Mixture Model & 1 & 3x64 & 1 \\
    Frey Faces & 1 & [512, 1024, 512] & 0 $\pm$ 12 \\
    Double Pendulum & 1 & [64, 128, 64] & 1 \\
    T Cell CyTOF & 2 & 2x256, 3x256, 2x256 & 80 $\pm$ 20 \\
    hESC scRNA-seq & 3 & 2x128, 3x128, 2x128 & Diffusion \\
    \bottomrule
    \end{tabular}
    \end{sc}
    \end{small}
    \end{center}
    \vskip -0.1in
    \label{tab:training}
\end{table}

\section{DyMoN Details}
\textbf{DyMoN architecture:} One could envision many neural network architectures which fit the schema of the DyMoN that we propose. In this paper, we use three alternative models; these are shown in Figure~\ref{fig:architectures}. The first architecture generates transitions on the ambient space, most suitable to data of low dimensionality. The other two architectures generate transitions on the latent space of an autoencoder, which learns a latent space which, due to gradient descent propagating through the encoder and decoder, learns a latent space more suited to generating transitions than either the ambient space or latent spaces learned by other dimensionality reduction methods. Additionally, the generation of points in a sampled trajectory can be passed through the encoder and decoder in order to prevent accumulation of error over the course of many samples.

\textbf{Training the DyMoN:} We train DyMoN using leaky ReLU activations on the hidden nodes and linear activations on the residual output nodes. Stochastic DyMoNs are trained with Gaussian noise inputs, and all DyMoNs are trained with Gaussian corruption noise. To compute the MMD loss, we use a multi-scale Gaussian kernel \cite{bousmalis2016domain} with 19 bandwidths ranging from $\num{1e-6}$ to $\num{1e6}$, evenly spaced on a log scale. The kernel is computed separately for each bandwidth and then MMD is computed on the sum of the kernels. The Adam optimizer~\cite{kingma2014adam} is used for stochastic gradient descent in all cases.



\textbf{Training time:} Table~\ref{tab:runtimes} shows DyMoN training times for the datasets considered.

\begin{table}[t]
    \centering
    \caption{Training time for empirical tests. All networks were trained with 2617MB of RAM on a NVIDIA Titan X Pascal GPU.}
    \vskip 0.15in
    \begin{center}
    \begin{small}
    \begin{sc}
    \begin{tabular}{lcr}
    \toprule
    Data set & Epochs & Time (min) \\
    \midrule
    Pendulum & 500 & 3.7 \\
    Teapot & 1000 & 8 \\
    Mixture Model & 600 & 9 \\
    Frey Faces & 1800 & 698 \\
    Double Pendulum & 1400 & 1801 \\
    \bottomrule
    \end{tabular}
    \end{sc}
    \end{small}
    \end{center}
    \vskip -0.1in
    \label{tab:runtimes}
\end{table}

\section{Methods comparison}

\begin{figure}[!h]
    \centering
    \includegraphics[width=\linewidth]{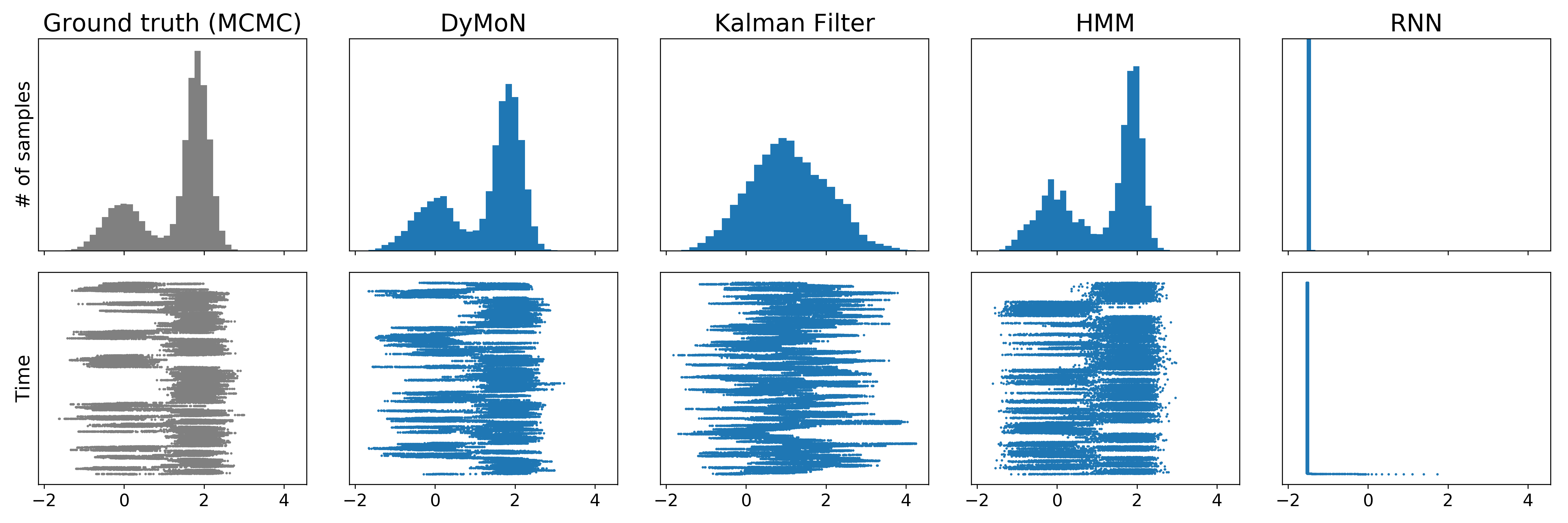}
    \caption{Chain of samples (top) and marginal distribution (bottom) drawn from Markov-Chain Monte Carlo (MCMC), DyMoN, Recurrent Neural Network (RNN), Hidden Markov Model (HMM) and Kalman Filter (KF) when trained on the Gaussian Mixture Model.}
    \label{fig:gmm_method_comparison}
\end{figure}

\begin{figure}
    \centering
    \includegraphics[width=\linewidth]{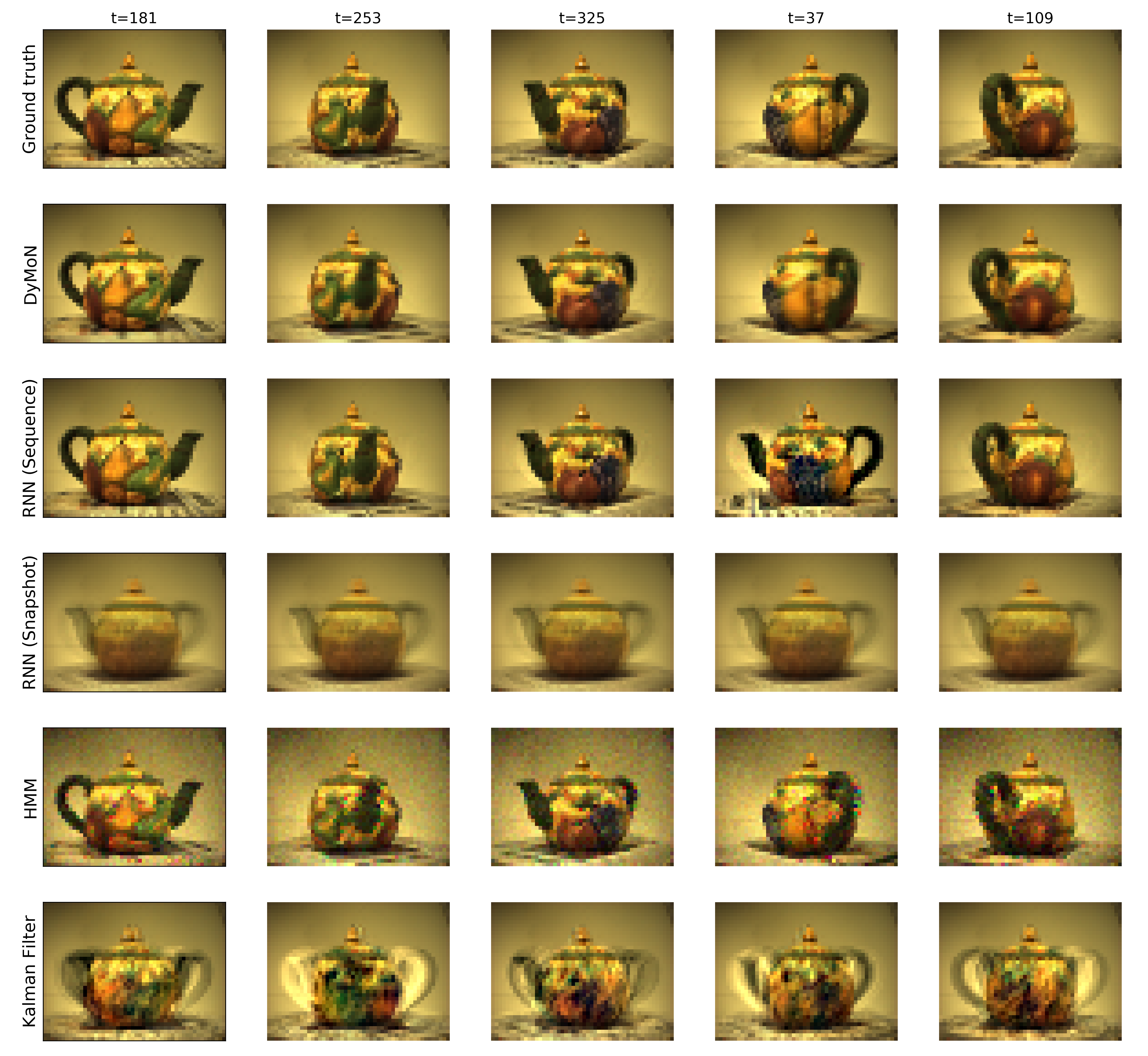}
    \caption{Examples of samples drawn from DyMoN, Recurrent Neural Network (RNN) given either a full sequence of inputs (Sequence) or a single input padded by zeroes (Snapshot), Hidden Markov Model (HMM) and Kalman Filter (KF) when trained on the teapot data.}
    \label{fig:teapot_method_comparison}
\end{figure}

Competing methods were trained on the same dataset as for DyMoN, and with similar architecture where possible. We used Hidden Markov Models provided by the \texttt{hmmlearn} Python package and Kalman Filters provided by the \texttt{pykalman} package, in both cases learning all parameters by Expectation-Maximization (EM). The EM algorithm was run with default parameters. Recurrent Neural Networks were trained with the same number of hidden layers and convolutions as the DyMoN, except in the teapot example where we found performance improved by adding an additional two hidden layers of size 64. Performance of the competing methods using Earth Mover's Distance (EMD) on the GMM example is shown in Figure~\ref{fig:gmm_method_comparison} and examples of generated frames from the teapot data are shown in Figure~\ref{fig:teapot_method_comparison}. 

\textbf{Pointwise sample comparison to other methods:} We train the a DyMoN with 2 convolutional layers with 5x5 filters and 2x2 max pooling, a single fully connected layer of 3 hidden nodes, and 2 deconvolutional layers on a video of a rotating teapot \cite{weinberger2004learning}. We compare the performance of RNNs, HMMs and KFs to DyMoN on this higher-dimensional example. We produce a single transition from each frame in the time series (giving a chain of prior states to the HMM, RNN and KF in order to facilitate their estimates of the current state) and measure the mean squared error between the produced output and the true frame, ten time points after the current state. Due to the relatively little training data available (400 samples), both classical statistical learning methods (HMM, KF) have difficulty generating accurate samples from this system given a single snapshot. RNNs perform well on the large majority of frames when given a sequence as input, but fail entirely when given only a single frame as input on inference, making them inherently unsuitable to inference in snapshot biological systems where sequential input is not available. Additionally, DyMoN is able to generate images faster than all other methods, and is only slower in training than the HMM, which produces significantly lower quality images. Examples of images produced by each method are shown in Figure~\ref{fig:teapot_method_comparison}.

\begin{table*}[b]
    \centering
    \caption{Performance of various learning methods for dynamical systems on generating frames from the teapot dataset.}
    \vskip 0.15in
    \begin{center}
    \begin{small}
    \begin{sc}
    \begin{tabular}{lccc}
    \toprule
                        & MSE               & Training (CPU s) & Inference (CPU s / forward pass) \\
    \midrule
        DyMoN           & $1.8 \pm 0.01$    & 8 min     & 1.2 s \\
        RNN (sequence)  & $2.3 \pm 0.20$    & 8.7 h     & 3.0 s \\
        RNN (snapshot)  & $9.1 \pm 0.04$    & ---       & 2.9 s \\
        KF              & $10.2 \pm 0.07$   & 300 h     & 175 s \\
        HMM             & $5.4 \pm 0.06$    & 28 s      & 192 s \\
    \bottomrule
    \end{tabular}
    \end{sc}
    \end{small}
    \end{center}
    \vskip -0.1in
    \label{tab:teapot_performance}
\end{table*}

\end{appendices}